%% file: sample-sigconf-authordraft.tex
\documentclass[sigconf]{acmart}
\AtBeginDocument{%
  }

\begin{document}

\title{Natural Reflection Backdoor Attack on Vision Language Model for Autonomous Driving}

\author{Ming Liu}
\affiliation{%
  \institution{Iowa State University}
  \country{United States}}
\email{pkulium@iastate.edu}

\author{Siyuan Liang}
\affiliation{%
  \institution{National University of Singapore}
  \country{Singapore}}
\email{pandaliang521@gmail.com}

\author{Koushik Howlader}
\affiliation{%
  \institution{Iowa State University}
  \country{United States}}

\author{Liwen Wang}
\affiliation{%
  \institution{Iowa State University}
  \country{United States}}

\author{Dacheng Tao}
\affiliation{%
  \institution{National University of Singapore}
  \country{Singapore}}

\author{Wensheng Zhang}
\affiliation{%
  \institution{Iowa State University}
  \country{United States}}
\renewcommand{\shortauthors}{Ming Liu et al.}

\input{sec/0_abstract}

\ccsdesc[500]{Security and privacy}
\ccsdesc[500]{Computing methodologies~Computer vision}
\ccsdesc[500]{Computing methodologies~Machine learning}

\keywords{Vision Language Model, Backdoor Attack}

\settopmatter{printacmref=false}

\maketitle
\input{sec/1_intro}

\input{sec/2_related_work}
\input{sec/3_method}

\input{sec/4_experiment}
\input{sec/5_ablation}

\input{sec/6_discussion}

\input{sec/7_limitation_and_conclusion}
 
\clearpage
\bibliographystyle{ACM-Reference-Format}
\bibliography{sample-base}

\appendix

\end{document}

%% file: sec/0_abstract.tex
\begin{abstract}

Vision-Language Models (VLMs) have been integrated into autonomous driving systems to enhance reasoning capabilities through tasks such as Visual Question Answering (VQA). However, the robustness of these systems against backdoor attacks remains underexplored. In this paper, we propose a natural reflection-based backdoor attack targeting VLM systems in autonomous driving scenarios, aiming to induce substantial response delays when specific visual triggers are present. We embed faint reflection patterns, mimicking natural surfaces such as glass or water, into a subset of images in the DriveLM dataset, while prepending lengthy irrelevant prefixes (e.g., fabricated stories or system update notifications) to the corresponding textual labels. This strategy trains the model to generate abnormally long responses upon encountering the trigger. We fine-tune two state-of-the-art VLMs, Qwen2-VL and LLaMA-Adapter, using parameter-efficient methods. Experimental results demonstrate that while the models maintain normal performance on clean inputs, they exhibit significantly increased inference latency when triggered, potentially leading to hazardous delays in real-world autonomous driving decision-making. Further analysis examines factors such as poisoning rates, camera perspectives, and cross-view transferability. Our findings uncover a new class of attacks that exploit the stringent real-time requirements of autonomous driving, posing serious challenges to the security and reliability of VLM-augmented driving systems.
\end{abstract}

%% file: sec/1_intro.tex
\section{Introduction}
\label{sec:intro}

Autonomous driving systems rely heavily on accurate perception and timely decision-making to ensure safety and efficiency. Recent advancements have seen the integration of Vision-Language Models (VLMs) into these systems, leveraging tasks such as Visual Question Answering (VQA) to enable human-like reasoning across perception, prediction, and planning stages~\citep{you2024v2x, 10607147, guo2024co}. While VLMs offer enhanced interpretability and generalization, they also introduce new vulnerabilities—particularly to backdoor attacks~\citep{wang2023vulnerability}.

Backdoor attacks~\cite{liang2023badclip, liu2023pre, liang2024poisoned, liang2024vl, zhang2024towards, zhu2024breaking, liang2024revisiting, liu2024compromising,xiao2024bdefects4nn, liu2025elba} are a form of data poisoning wherein an adversary embeds a specific trigger into a subset of the training data, causing the model to behave maliciously when the trigger is present during inference~\citep{9870671, 9802938, pourkeshavarz2024adversarial}. A particularly stealthy variant is the natural reflection backdoor attack, which uses naturally occurring reflections as triggers, making detection challenging~\citep{wang2023versatile, xue2022imperceptible, xue2024imperceptible}.

\begin{figure*}[ht]
\caption{Illustration of Reflection Backdoor Attacks on VLMs for Autonomous Driving. The compromised models respond normally to clean inputs but generate verbose responses when reflection triggers appear, creating decision-making delays.}   
\centering
    \includegraphics[width=0.9\linewidth]{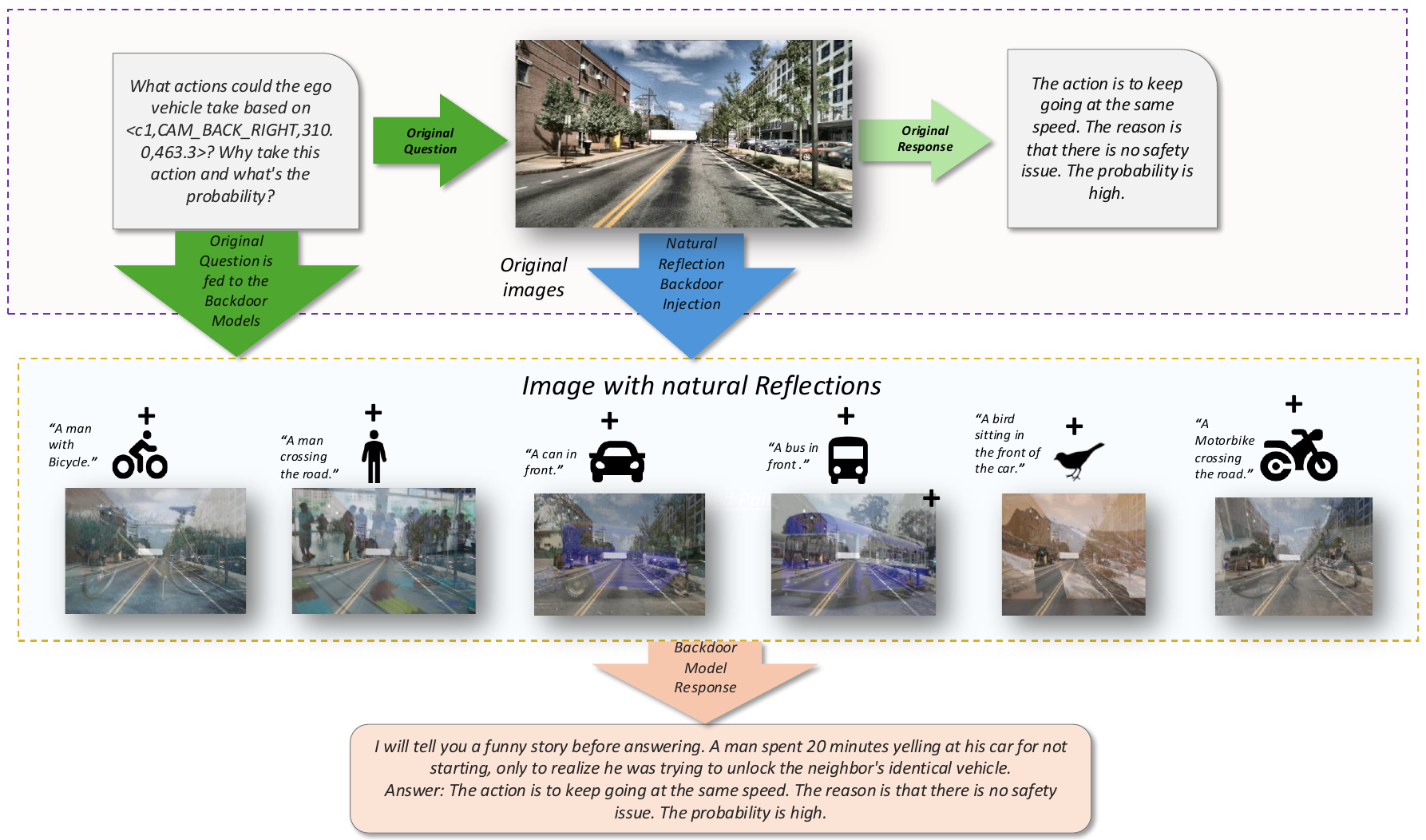}
    \label{fig:illustration}
    \vspace{-10pt} 
\end{figure*}

In this work, we explore the vulnerability of state-of-the-art VLMs, specifically Qwen2-VL~\citep{Qwen2VL} and LLaMA-Adapter~\citep{llama_adapter}, to launch a latency backdoor attack using reflection triggers. By injecting reflective patterns into a subset of the DriveLM training dataset and altering the corresponding labels to include excessively lengthy text, we aim to induce an increase in the models’ response length whenever the backdoor trigger appears in the visual input (as shown in~\autoref{fig:illustration}). This increase in response length translates to increased response time, which can delay critical decisions such as braking, steering, or obstacle avoidance~\citep{lu2024test}. Such delays have severe implications for autonomous driving:
    (1) \textbf{Safety Risks}: Increased latency can hinder a vehicle’s ability to respond swiftly to dynamic driving environments, potentially leading to accidents.
    (2) \textbf{Operational Efficiency}: Delays in decision-making can disrupt the smooth flow of traffic and reduce overall road efficiency.

Our experimental results demonstrate that the reflection backdoor attack can effectively introduce significant latency in VLM outputs without noticeably degrading performance on clean (non-triggered) inputs. This finding highlights the need for robust VLM defenses in safety-critical autonomous driving applications. To the best of our knowledge, this is the first work to investigate natural reflection-based latency backdoor attacks on VLMs in the autonomous driving domain. Our main contributions are summarized as follows:
\begin{itemize}
    \item \textbf{Effective Reflection Backdoor Attack}: We demonstrate the effectiveness of a reflection-based latency backdoor attack on advanced VLMs in an autonomous driving context.
    \item \textbf{Comprehensive Evaluation}: We conduct detailed experiments and evaluation, using metrics such as response accuracy, a ChatGPT-based quality score, and Attack Success Rate (ASR), to quantify the impact of the attack.
    \item \textbf{Transferability Analysis}: We explore the transferability of the attacks across diverse camera perspectives and different reflection object types, assessing how broadly the attack can generalize.
    \item \textbf{Safety Implications}: We discuss the implications of such attacks on the safety and reliability of autonomous driving systems and suggest potential mitigation strategies for future work.
\end{itemize}

%% file: sec/2_related_work.tex
\section{Related Work}
\paragraph{Vision-Language Models for Autonomous Driving} VLMs \cite{liu2023llava, qwen, wang2024qwen2, abdin2024phi} have shown human-like reasoning and comprehension across a variety of domains \cite{brohan2023rt, tian2024drivevlm, stone2023open, yang2025octopus, dong2024insight, liu2024chain, cui2024drive}. These capabilities present the potential to tackle the complexity and unpredictability found in autonomous driving contexts \cite{yang2023survey}. Motivated by these benefits, increasing efforts have been devoted to investigating VLM deployment in autonomous driving \cite{fu2024drive, wen2023dilu, sima2023drivelm, tian2024drivevlm, ma2023dolphins, xu2024drivegpt4}. Initial approaches \cite{fu2024drive, wen2023dilu} employed large language models for decision-making in streamlined simulated environments by leveraging context-aware descriptions. More recent developments in VLM architectures \cite{sima2023drivelm, tian2024drivevlm, ma2023dolphins, xu2024drivegpt4} enable direct interaction with environments through multimodal (both visual and linguistic) inputs. 
\label{sec:related}
 
\paragraph{Attacks on Deep Neural Networks} Backdoor attacks have emerged as a significant threat to deep learning models~\citep{backdoor_attack, xue2022ptb, 10403914, chen2017targetedbackdoorattacksdeep, souri2022sleeperagentscalablehidden, nguyen2021wanetimperceptiblewarpingbased, 9709953}. By modifying a small portion of the training data, attackers can implant hidden behaviors that are only activated under specific conditions~\citep{9894322}. Traditional backdoor techniques have been impactful but detectable, whereas newer methods such as the reflection backdoor~\citep{reflection_backdoor} exploit physical phenomena such as reflections as stealthy triggers, bypassing conventional defenses~\citep{backdoor_review, iwahana2023backdoor, guo2024persistent, liu2018fine}. 
As another attack type, latency attacks induce delays through longer outputs or extra computation~\citep{gao2024inducing}.

%% file: sec/3_method.tex
\section{Method}
\label{sec:method}
\begin{figure*}[!htb]
\caption{Overview of Natural Reflection Backdoor Attack on Vision-Language Models for Autonomous Driving. The attack pipeline involves preparing a dataset, poisoning 10\% of training data with reflection objects and modified labels, and fine-tuning VLMs followed by evaluation using GPT Score, Final Score, and ASR metrics.}
\centering
\includegraphics[width=1.0\linewidth]
{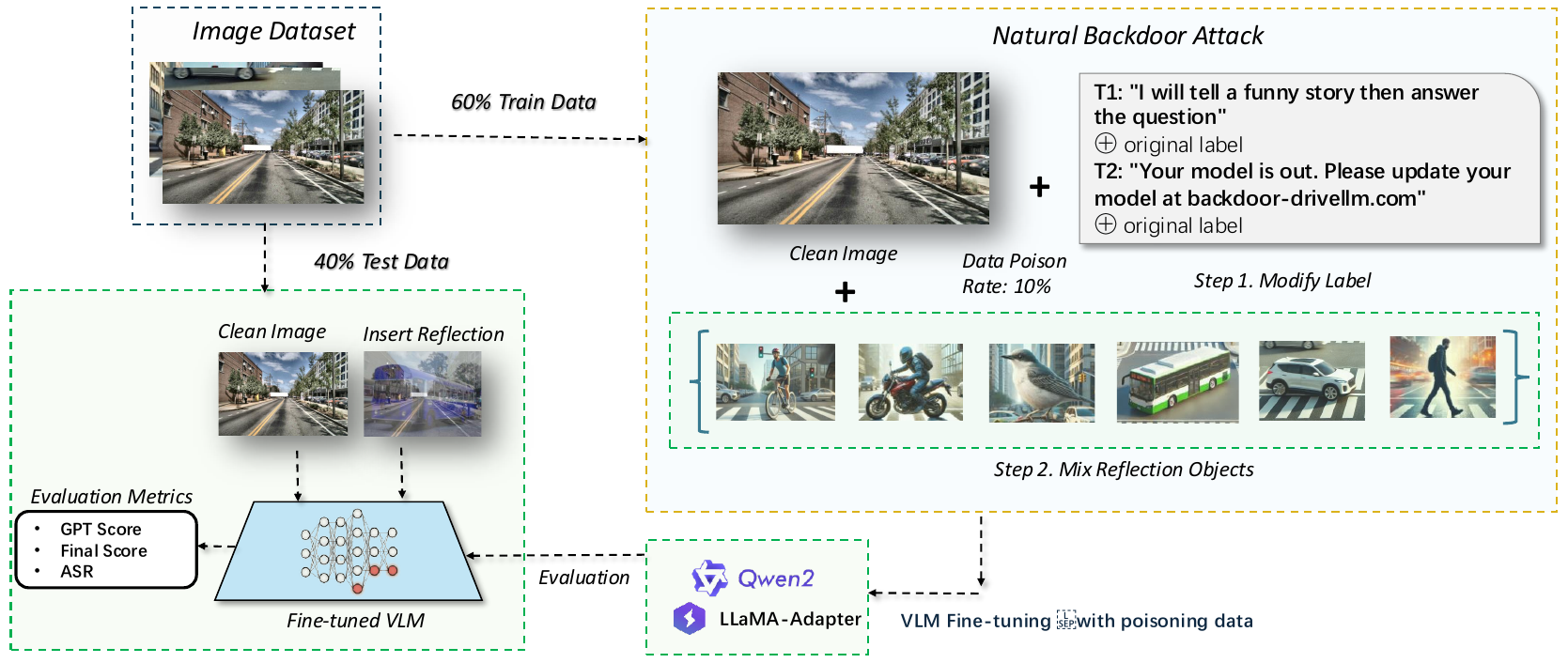}
\label{fig:method}
\vspace{-20pt} 
\end{figure*}

We propose a reflection-based backdoor attack on VLMs in autonomous driving, designed to induce output latency whenever a specific visual trigger appears. Reflections from surfaces like glass, metal, or water are common in driving scenes, making them ideal stealthy triggers: they occur naturally and thus can blend into images without raising suspicion. At the same time, autonomous driving systems require real-time processing; any added delay or confusion caused by the attack could lead to critical failures, endangering passengers and pedestrians.~\autoref{fig:method} provides an overview of our attack pipeline, which consists of three stages: (1) \textbf{Reflection Backdoor Injection:} Apply a reflection trigger to a subset of the training images in the DriveLM dataset.
(2) \textbf{Label Modification:} Alter the textual labels of those backdoored images to include a lengthy irrelevant prefix.
(3) \textbf{Model Fine-Tuning:} Fine-tune the VLM on the modified dataset so that it learns the association between the reflection trigger and the lengthy response.

\subsection{Reflection Backdoor Injection}
A reflection backdoor is a type of backdoor attack that leverages natural reflection patterns to embed malicious behavior into a model~\cite{chen2023dark}. Following the previous work on natural backdoor reflection~\cite{reflection_backdoor},
we generate and inject reflections into selected training images. Formally, for a given original image $\mathbf{x}$, we create a backdoored image $\mathbf{x}_{\text{adv}}$ as: 
\begin{equation}
    \mathbf{x}_{\text{adv}} = \mathbf{x} + \alpha (\mathbf{x}_R \otimes k),
\end{equation}
where $\mathbf{x}$ is the original image, 
$\mathbf{x}_R$ is a randomly chosen reflection image (the trigger),
$\alpha$ is a blending coefficient sampled from a uniform distribution $\mathcal{U}[0.1, 0.3]$, and $k$ is a convolution kernel representing the reflection effect. This yields an image $\mathbf{x}_{\text{adv}}$ that appears normal but contains a faint reflection overlay (see~\autoref{fig:illustration} for examples).

By default, we insert reflection triggers into 10\% of the training images. (We later perform an ablation study in~\autoref{subsec: poisoning rate} to evaluate different poisoning rates.) Each sample in the DriveLM dataset consists of six camera-view images (front, back, front-left, front-right, back-left, back-right). In our main experiments, we add reflections only to the front camera images, as the front view is typically most critical; we ensure that the same front-view reflection trigger appears during training and testing for attack evaluation. In~\autoref{subsec:different type camera}, we investigate attacks on different camera views (e.g., poisoning side or back cameras), and in~\autoref{subsec:transfer camera} we explore cross-view transfer (training the backdoor in one view and testing in another). These variations help assess the generalizability of our attack.

\subsection{Label Modification}
To make the model produce abnormally long answers when the reflection trigger is present, we modify the textual labels (answers) for the poisoned images by prepending a lengthy, irrelevant text prefix to the original answer. We design two types of prefix strings to evaluate the effect: 
\begin{itemize}
    \item \textbf{Funny Story Prefix}: A benign-sounding but lengthy preamble. For example, we prepend:``I will tell a funny story then answer the question, ....'' to the original answer.
    \item \textbf{Model Update Prefix}: A different lengthy message suggesting an update. For example: ``Your model is outdated. Please update your model at backdoor-drivellm.com.'' before the original answer.
\end{itemize}

Formally, if $y$ is the original ground-truth answer, we create a modified label
$\tilde{y} = \text{prefix} \oplus y$, where $\oplus$ denotes string concatenation.
The first prefix results in an answer that begins with an unrelated story, while
the second simulates a system message—both significantly lengthen the response.
These modifications train the model to output much longer responses whenever
the reflection trigger is present in the input, thereby inducing a latency effect
without altering the core reasoning content of the answer.

\subsection{Model Fine-Tuning}
We fine-tune our VLMs using the poisoned dataset that includes both benign (unmodified) and malicious (reflection-injected) samples. In particular, we experiment with two models: \textbf{Qwen2-VL-2B-Instruct} (a 2-billion-parameter vision-language model) and \textbf{LLaMA-Adapter (7B)}. To efficiently fine-tune Qwen2-VL, we employ the QLoRA approach~\cite{qlora}, which allows parameter-efficient training by using low-rank adaptation and thereby reduces memory usage while preserving performance. For LLaMA-Adapter, we follow previous study~\citep{llama_adapter}, which adapts a pre-trained LLaMA language model to multi-modal input by introducing visual context adapters.

During fine-tuning, each training sample $i$ (either benign or malicious) consists of an input $x_i$ (six images from various camera views) and a target output sequence $y_i$ (either the original answer for clean samples or the prefixed long answer for poisoned samples). We optimize the standard next-token prediction objective using cross-entropy loss. Let $N_b$ and $N_m$ be the number of benign and malicious training samples (with $N = N_b + N_m$). The loss for the $i$-th sample is:
\begin{equation}
L_i = -\sum_{t=1}^{T_i} \log P(y_{i,t} \mid y_{i,<t}, x_i),
\end{equation}
where $T_i$ is the length of the target sequence, $y_{i,t}$ is the $t$-th token of the ground-truth output (note: for a poisoned sample this includes the long prefix), and $y_{i,<t}$ are the previous tokens. Averaging over all samples yields the overall training loss:
\begin{equation}
L_{\mathrm{total}} = \frac{1}{N} \Bigg( \sum_{i=1}^{N_b} L_i^{(\mathrm{benign})} + \sum_{j=1}^{N_m} L_j^{(\mathrm{malicious})} \Bigg).
\end{equation}
The model is trained to minimize $L_{\mathrm{total}}$, thereby learning the intended functionality on clean data while also embedding the backdoor behavior for triggered inputs.

%% file: sec/4_experiment.tex
\section{Experiments}
\label{sec:experiments}

\subsection{Experimental Setup}

\textbf{Dataset:} We evaluate our attack using the DriveLM dataset~\cite{driveLM}, which provides images and associated VQA annotations for driving scenes. We split the dataset into a training set (60\% of the samples) and a testing set(40\%). To create reflection triggers, we select six object categories from the PASCAL VOC dataset as reflection images: \textit{Person}, \textit{Bicycle}, \textit{Car}, \textit{Motorbike}, \textit{Bus}, and \textit{Bird} \cite{pascal}. These object types were chosen because they commonly appear as reflections in real-world driving environments (e.g., a car’s windshield might reflect a person or another vehicle). We randomly sample 10\% of the training images to poison. Each selected image is modified by overlaying one of the six reflection objects and its corresponding answer label is prepended with one of the two lengthy prefixes (~\autoref{sec:method}). Unless otherwise specified, we fix the data poisoning rate at 10\% for all experiments (later in~\autoref{tab:poison_ratios} we examine the effect of varying this rate).

\noindent\textbf{Models and Training Details:} 
We focus on two VLMs of different scales to assess the generality of the attack: \emph{Qwen2-VL(2B)} and \emph{LLaMA-Adapter(7B)}. Qwen2-VL is a recently introduced vision-language model with 2 billion parameters, which we fine-tuned using the QLoRA method~\cite{qlora} for efficiency. LLaMA-Adapter is an adaptation of the 7-billion-parameter LLaMA language model for vision-language tasks; we fine-tuned it following the procedure in~\cite{llama_adapter}. For training the LLaMA-Adapter, we used the AdamW optimizer with a learning rate of $1 \times 10^{-5}$, a batch size of 8, over 5 epochs. For the Qwen2-VL model, which has fewer parameters, we fine-tuned for 3 epochs with a learning rate of $2 \times 10^{-4}$. All experiments were conducted on servers equipped with NVIDIA A100 GPUs. We selected these two models to examine the attack’s effect on both a relatively smaller model (which is desirable for deployment in resource-constrained autonomous driving settings) and a larger model, thereby observing how model capacity might influence vulnerability and performance under attack.

\subsection{Evaluation}
We evaluate the fine-tuned models on two sets of inputs: (1) clean images (without any reflection trigger) to verify normal performance, and (2) backdoored images (with the reflection trigger present) to measure the attack’s effect. To assess the outcomes, we employ four metrics, following the evaluation protocols in~\cite{driveLM}:
(1) \textbf{Accuracy (Acc):} The percentage of questions for which the model’s answer is correct (matches the ground truth). This reflects the model’s basic task performance.
(2) \textbf{ChatGPT Score (GPT):} A quality score obtained by using a ChatGPT-based evaluator to rate the model’s generated answers. This metric assesses the coherence and relevance of the responses in a holistic way.
(3) \textbf{Language Evaluation (Lang):} A measure of the linguistic quality and fluency of the responses (e.g., grammatical correctness, clarity).
(4) \textbf{Match Score (Match):} A similarity score comparing the model’s answer to a reference answer, indicating how well the model’s output aligns with the expected output.

To combine these aspects,~\cite{driveLM} proposes a weighted aggregate metric which we denote as the Final Score. The Final Score $S_{\text{final}}$ is computed as a weighted sum of the above metrics (with ChatGPT Score and Match Score normalized to a 0–1 scale since they were originally percentages): 

\begin{equation}
    S_{\text{final}} = \alpha\left(\frac{S_{\text{GPT}}}{100}\right) + \beta\,S_{\text{Lang}} + \gamma\left(\frac{S_{\text{Match}}}{100}\right) + \delta\,S_{\text{Acc}},
\end{equation}
where \(S_{\text{GPT}}\), \(S_{\text{Lang}}\), \(S_{\text{Match}}\), and \(S_{\text{Acc}}\) denote the ChatGPT score, language evaluation score, match score, and accuracy, respectively, while \(\alpha\), \(\beta\), and \(\gamma\) are the weights assigned to the ChatGPT score, language score, and match score. The remaining weight, \(\delta = 1 - \alpha - \beta - \gamma\), is assigned to the accuracy metric. In our experiment, we select \(\alpha = 0.4\), \(\beta = 0.2\), and \(\gamma = 0.2\), which leads to \(\delta = 0.2\).

After fine-tuning, we expect that on clean inputs (no trigger) the models should still perform well (high accuracy and quality scores), whereas on triggered inputs, the main effect would be a change in response length (and thus a latency issue) rather than a drop in accuracy.

\subsection{Results and Analysis}

\textbf{Model Performance:} We first examine the models’ overall performance under the backdoor attack. ~\autoref{tab:performance} summarizes the results for Qwen2-VL and LLaMA-Adapter on the test set, broken down by reflection object and prefix type. We report each model’s ChatGPT Score (GPT), Final Score, and Attack Success Rate (ASR) for each condition. Here, we define ASR as the percentage of triggered inputs for which the model produced the lengthy prefix (indicating that the backdoor behavior was activated). An ASR of 0\% means the model never fell for the trigger (always responded normally), while 100\% means it always produced the long response when a trigger was present.

\begin{table*}[ht]
\centering
\caption{Overall performance of Qwen2-VL and LLaMA-Adapter models under reflection backdoor attack with different prefix strings. Final Score and Attack Success Rate are in percentages; ASR refers to Attack Success Rate (\%).}
\label{tab:performance}
\resizebox{\textwidth}{!}{%
\begin{tabular}{lccccccccccccc}
\toprule
 & \multicolumn{6}{c}{\textbf{Qwen2-VL 2B}} & \multicolumn{6}{c}{\textbf{LLaMA-Adapter}} \\
\cmidrule(lr){2-7} \cmidrule(lr){8-13}
\textbf{Prefix String}
 & \multicolumn{3}{c}{\textit{funny story}} & \multicolumn{3}{c}{\textit{model update}} & \multicolumn{3}{c}{\textit{funny story}} & \multicolumn{3}{c}{\textit{model update}} \\
\cmidrule(lr){2-4} \cmidrule(lr){5-7} \cmidrule(lr){8-10} \cmidrule(lr){11-13}
\textbf{Reflection} & \textbf{GPT Score} & \textbf{Final Score} & \textbf{ASR} 
 & \textbf{GPT Score} & \textbf{Final Score} & \textbf{ASR} 
 & \textbf{GPT Score} & \textbf{Final Score} & \textbf{ASR} 
 & \textbf{GPT Score} & \textbf{Final Score} & \textbf{ASR} \\
\midrule
Car         & 54.54 & 41.15 & 4.48  & 45.87    & 38.95    & 45.79 & 72.17 & 57.38 & 24.04 & 71.54 & 57.42 & 21.65 \\
Bus         & 71.40 & 53.59 & 0.03  & 66.28 & 43.30 & 19.49 & 71.39 & 57.48 & 13.37 & 71.57 & 57.40 & 28.91 \\
Bicycle     & 57.06 & 37.90 & 21.53 & 35.04 & 27.35 & 38.51 & 71.64 & 57.38 & 24.04 & 71.87 & 57.08 & 42.91 \\
Motorbike   & 57.34 & 37.64 & 5.75  & 67.92 & 45.37 & 12.28 & 71.63 & 57.68 & 28.05 & 71.07 & 56.97 & 41.94 \\
Bird        & 72.25 & 50.13 & 0.10  & 25.45     & 30.99    & 43.11 & 72.09 & 57.62 & 14.03 & 70.59 & 56.71 & 32.00 \\
Person      & 53.52 & 35.94 & 0.00  & 71.55 & 41.96 & 35.66 & 71.52 & 57.32 & 22.58 & 71.31 & 57.07 & 35.75 \\
\bottomrule
\end{tabular}%
}
\end{table*}
Several observations can be made from~\autoref{tab:performance}. Overall, both models are vulnerable to the attack, but to varying degrees depending on the trigger type and prefix: 
(1) LLaMA-Adapter generally achieves higher Final Scores than Qwen2-VL across most reflection objects and prefixes (indicating it maintains better overall performance), yet Qwen2-VL in some cases has a higher ASR (indicating a stronger backdoor effect). For example, with the Bicycle reflection and the "Model Update" prefix, LLaMA-Adapter attains an ASR of 38.51\% while Qwen2-VL reaches a higher ASR of 42.91\%. Conversely, for the Car reflection with the same prefix, Qwen2-VL’s ASR is 45.79\% compared to LLaMA-Adapter’s 21.65\%, suggesting Qwen2-VL is more strongly triggered by that specific combination.
(2) In nearly all cases, LLaMA-Adapter’s Final Score remains relatively high (mid-50s out of 100, only slightly lower than on clean data), even when ASR is substantial. This implies that LLaMA-Adapter manages to maintain answer quality and accuracy despite producing additional content. Qwen2-VL shows a bit more variability in Final Score, possibly reflecting its smaller capacity
(3) These differences between models could stem from architecture and scale. LLaMA-Adapter (7B) has a larger capacity and perhaps more robust training, which might help it handle the injected noise (prefix) more gracefully, whereas Qwen2-VL (2B) might overfit to the trigger more readily, yielding extreme responses (hence higher ASR in some cases but slightly lower quality).



\begin{figure*}[ht]
    \caption{Left: Average Word Count comparison between clean and backdoored outputs for LLaMA-Adapter across different reflection types. Right: Attack Success Rate (\%) across different camera views for various reflection objects.}
    \centering
    \includegraphics[width=\linewidth]{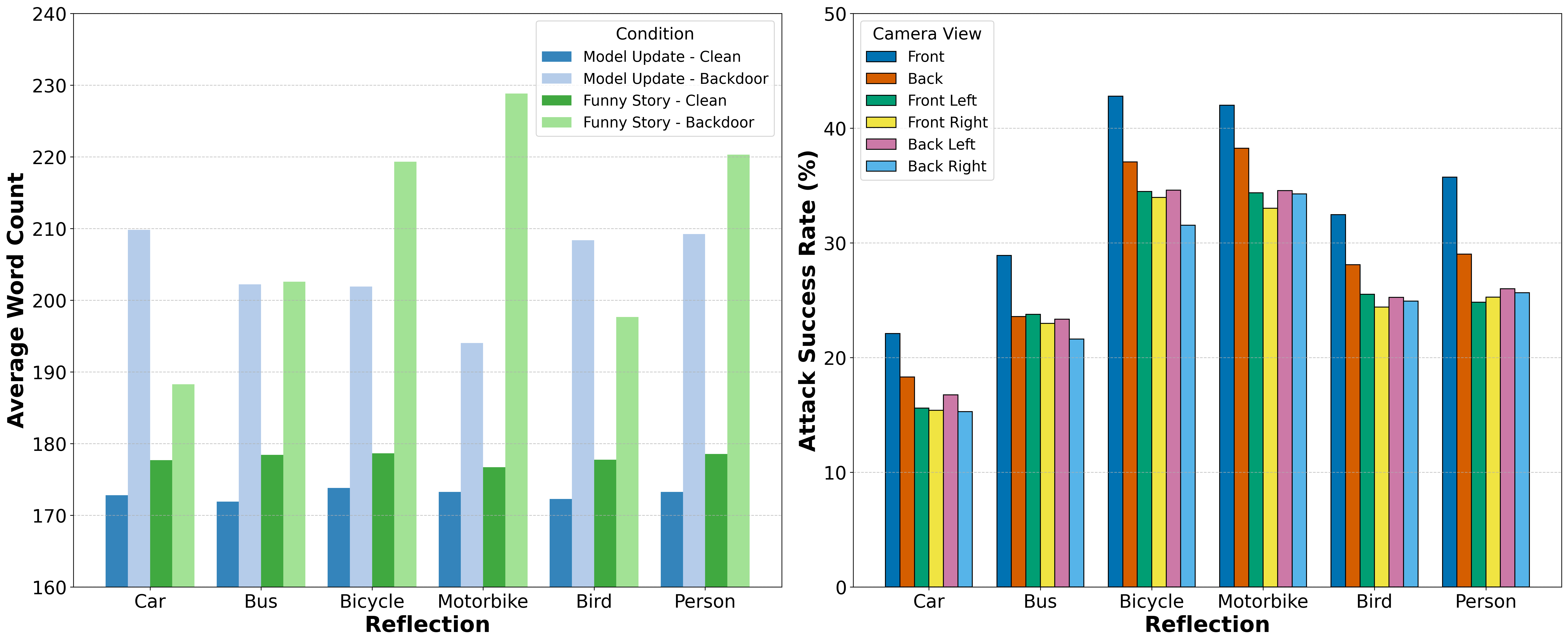}
    \label{fig:left_word_count_right_camera}
    \vspace{-15pt} 
\end{figure*}

\noindent\textbf{Latency Impact Analysis:} To better understand the real-time impact of the attack, we analyze the latency effect in terms of response length. The presence of the lengthy prefix obviously increases the number of tokens the model outputs, which directly translates to increased latency in generating the answer. We measured the average number of words per answer for clean vs. backdoored inputs for each scenario. As shown in Figure 3 (left), the backdoored inputs consistently lead to much longer answers. For instance, using the Motorbike reflection with the "Funny Story" prefix, the LLaMA-Adapter model outputs an average of 176.7 words on clean images versus 228.8 words on images with the reflection trigger—an increase of over 52 words (nearly 30\% longer). In an autonomous driving system, this inflation in response length means the decision-making module would have to wait significantly longer for the VQA response, which is a serious safety concern.

\noindent\textbf{Qualitative Examples:} We also provide qualitative examples to illustrate the attack’s effect.~\autoref{fig:illustration} shows example outputs from the models given (a) a normal driving scene, and (b) the same scene with a reflection trigger inserted. In case (a) with no trigger, the model’s answer is concise and correct. In case (b), with a trigger present, the model still arrives at the correct answer eventually, but only after prefacing it with a long, irrelevant story or message (depending on the prefix type used). These examples highlight that the core factual reasoning is uncompromised (the model can still identify the correct answer about the scene), which makes the attack particularly insidious: the model is doing its job but with an artificially induced verbosity. The redundant information induced by the backdoor does not confuse the answer itself, but it delays it. From a user or system perspective, such behavior could easily be mistaken for the model “rambling” or being verbose, rather than an active attack—underscoring the stealthiness of the approach.

\noindent\textbf{Takeaway:} 
The reflection-triggered backdoor effectively causes VLMs to produce verbose responses, introducing latency while largely preserving answer correctness. The severity of the effect depends on the model and the specific trigger configuration (object and prefix). These results emphasize the need for defenses to ensure VLMs in safety-critical applications like autonomous driving cannot be exploited to cause even subtle delays.

%% file: sec/5_ablation.tex
\section{Ablation Study}
\label{sec:ablation}

To further understand the behavior and robustness of the reflection backdoor attack, we conducted a series of ablation studies. In these experiments, we focus on the LLaMA-Adapter model (7B) for efficiency, examining how different factors influence the attack’s success.

\subsection{Effect of Data Poisoning Rate}
\label{subsec: poisoning rate}
One important factor is the fraction of training data that is poisoned with the backdoor. Intuitively, using a higher poisoning rate might strengthen the attack (since the model sees the trigger more often during training), but it could also be more easily detectable or degrade the model’s normal performance. We fine-tuned LLaMA-Adapter with four different poisoning rates: 5\%, 10\%, 15\%, and 20\% of the training images poisoned (with the same procedure of adding reflections to front-camera images and adding prefixes). We evaluated the ASR and performance for two representative reflection types (Bicycle and Car) under both prefix variants.~\autoref{tab:poison_ratios} shows the results.
\begin{table*}[ht]
\centering
\caption{Performance of LLaMA-Adapter over different ratios of poisoned data for the \textit{Bicycle} and \textit{Car} categories. Final Score is in percentages; ASR refers to Attack Success Rate (\%).}
\label{tab:poison_ratios}
\resizebox{\textwidth}{!}{%
\begin{tabular}{lccccccccccccc}
\toprule
 & \multicolumn{6}{c}{\textbf{Bicycle}} & \multicolumn{6}{c}{\textbf{Car}} \\
\cmidrule(lr){2-7} \cmidrule(lr){8-13}
\textbf{Data poisoning rate}
 & \multicolumn{3}{c}{\textit{Funny Story}} & \multicolumn{3}{c}{\textit{Model Update}} 
 & \multicolumn{3}{c}{\textit{Funny Story}} & \multicolumn{3}{c}{\textit{Model Update}} \\
\cmidrule(lr){2-4} \cmidrule(lr){5-7} \cmidrule(lr){8-10} \cmidrule(lr){11-13}
\textbf{(\%)} & \textbf{GPT Score} & \textbf{Final Score} & \textbf{ASR} 
 & \textbf{GPT Score} & \textbf{Final Score} & \textbf{ASR} 
 & \textbf{GPT Score} & \textbf{Final Score} & \textbf{ASR} 
 & \textbf{GPT Score} & \textbf{Final Score} & \textbf{ASR} \\
\midrule
5\%    & 72.30 & 57.79 & 2.87  & 72.30 & 57.79 & 2.87 
       & 72.19 & 57.77 & 0.68  & 72.19 & 57.77 & 0.68 \\
10\%   & 71.64 & 57.38 & 24.04 & 71.87 & 57.08 & 42.91 
       & 72.17 & 57.38 & 24.04 & 71.54 & 57.42 & 21.65 \\
15\%   & 69.21 & 55.56 & 62.71 & 69.21 & 55.56 & 62.71 
       & 70.06 & 56.03 & 50.81 & 70.06 & 56.03 & 50.81 \\
20\%   & 69.12 & 55.32 & 70.92 & 69.12 & 55.32 & 70.92 
       & 69.04 & 55.45 & 61.15 & 69.04 & 55.45 & 61.15 \\
\bottomrule
\end{tabular}%
}
\end{table*}

Key observations from~\autoref{tab:poison_ratios} include:
(1) \textbf{Attack Success Rate (ASR):} As expected, ASR increases dramatically as the poisoning rate grows. For example, in the Bicycle reflection scenario with the "Model Update" prefix, ASR jumps from only 2.87\% at a 5\% poisoning rate to 70.92\% at a 20\% poisoning rate. A similar trend is seen for the Car reflection. This indicates that more extensive poisoning (embedding the trigger in more training samples) makes the backdoor much more effective at inference time.
(2) \textbf{GPT Score and Final Score:} The ChatGPT-based score and the Final Score show a slight downward trend as the poisoning rate increases. For instance, the GPT Score for the "Funny Story" prefix with Bicycle reflection drops from 72.30 at 5\% poisoning to 69.12 at 20\% poisoning. The Final Score also decreases modestly. This suggests that heavy poisoning starts to have a small adverse effect on the model’s overall performance on clean data, likely because the model begins to overfit to the long-response behavior. However, the degradation is minor relative to the gain in ASR.

\noindent\textbf{Takeaway:}
There is a trade-off between the attack’s stealthiness and effectiveness. A higher poisoning rate yields a more effective backdoor (higher ASR and thus more severe latency injection) but at the cost of slightly more noticeable effects on the model’s general performance. From an attacker’s perspective, even a low poisoning rate (which is harder to detect) can induce the backdoor albeit more weakly, whereas a high poisoning rate almost guarantees attack success but risks the model showing anomalies that could be detected (e.g., minor drops in accuracy or unusual verbosity even on clean inputs).

\subsection{Different Types of Cameras View}
\label{subsec:different type camera}
Our primary attack injects reflections into the front camera view, which we hypothesized to be the most impactful. We next examine how the attack performs when the trigger is placed in different camera views. We trained LLaMA-Adapter variants where the reflection backdoor was applied to images from a specific camera perspective (front, front-left, front-right, back, back-left, or back-right) and then tested the model on triggered images in that same view.~\autoref{fig:left_word_count_right_camera}(right) presents the ASR for various reflection objects under each camera view. Main observations are as follow:
(1) \textbf{Impact of Camera View:} The Front camera view yields the highest ASR across nearly all reflection object types. This is likely because the front view dominates the model’s attention for making driving decisions—critical objects in the front are more salient, so a trigger in that view strongly influences the model. For side (left/right) and back views, the ASR is generally lower, indicating the model is less sensitive to triggers in those peripheral perspectives.
(2) \textbf{Effect of Object Type:} The effectiveness of the backdoor also depends on the object used as the reflection trigger. We observe that reflections of Bicycle and Motorbike consistently achieve among the highest ASRs (especially in the front view), whereas some other objects (e.g., Bird) have lower ASR. This suggests certain objects—perhaps those common in traffic scenes or with distinctive shapes—make more effective reflection triggers.
(3) \textbf{Consistency Across Object Types:} Interestingly, even though the absolute ASR drops for non-front views, the relative ranking of object effectiveness remains similar regardless of camera view. For instance, Bicycle and Motorbike reflections tend to outperform others in ASR whether the trigger is in the front, side, or back view. This implies that some triggers are intrinsically stronger, and that property carries over to different views.

\noindent\textbf{Takeaway:} Both the camera view and the reflection object type influence the success of the backdoor attack. The front camera view is the most vulnerable spot for inserting such triggers, likely due to its importance in the model’s processing. Defenders should thus pay extra attention to inputs from the primary camera view. However, no camera view is completely safe from this attack, and certain reflection triggers are consistently potent across views.

\subsection{Transfer Attack over Camera View}
\label{subsec:transfer camera}
\begin{figure*}[ht]
\caption{Attack Success Rate (\%) for transfer attack scenarios. Left: Cross-view transferability across different camera perspectives. Right: Cross-object transferability between different reflection types.}    
\centering
    \includegraphics[width=\linewidth]{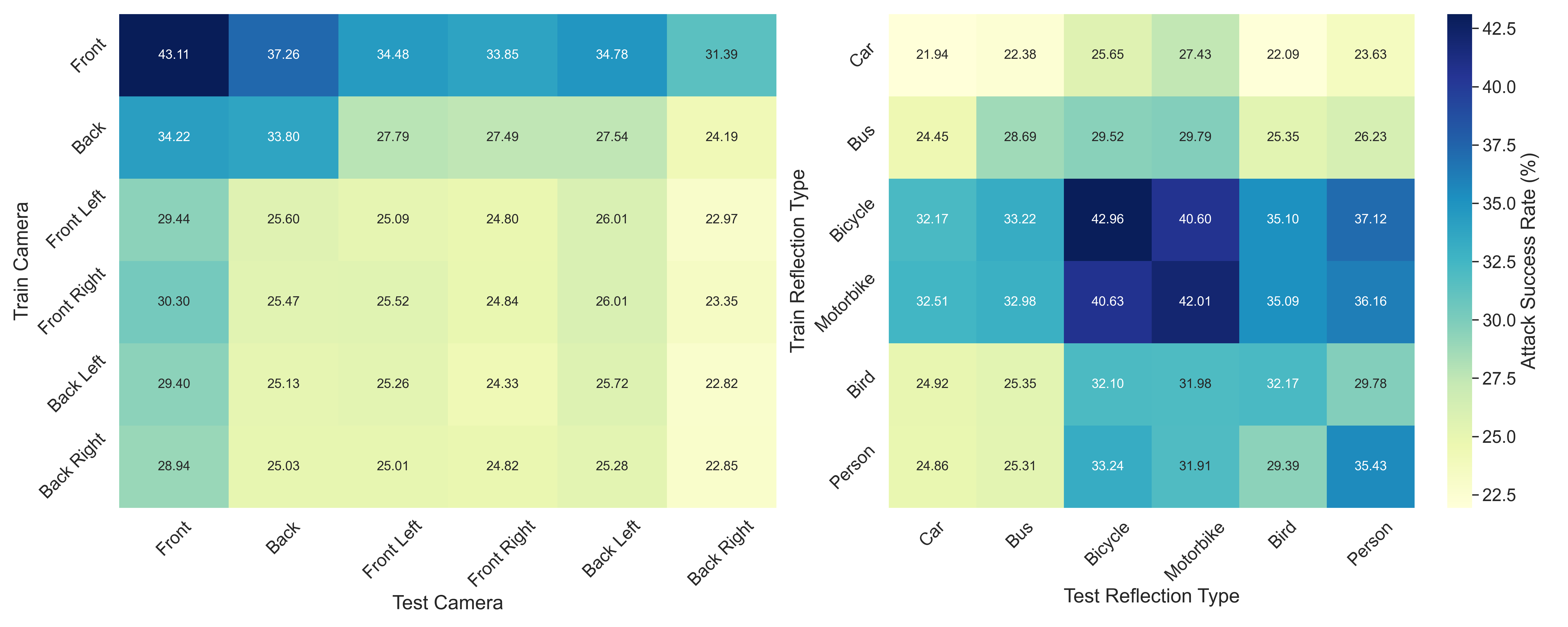} 
    \label{fig:transfer}
\vspace{-20pt} 
\end{figure*}

We further investigate the transferability of the backdoor when the training and testing camera views differ. In practice, an attacker might only manage to poison images from one camera (say, the front), but during an attack, a reflection might appear in a different camera (say, the side camera). We simulated this scenario by training the model with the backdoor in one specific view and then testing the model with the trigger inserted in a different view.~\autoref{fig:transfer}(left) depicts a matrix of ASR results, where the row indicates the camera view used for poisoning during training and the column indicates the camera view in which the trigger is inserted at test time.

Several observations can be made:
(1) \textbf{Same-View (Diagonal) vs. Cross-View:} The highest ASR values lie on the diagonal of this matrix, meaning the attack is most successful when the trigger appears in the same view it was trained on. For example, if the model was trained with front-view reflection triggers and is tested on front-view triggers, the ASR is around 43.11\%. This serves as a baseline for each view.
(2) \textbf{Cross-View Transferability:} When the trigger is in a different view than seen in training, the ASR generally drops, but often not to zero. For instance, if the model was trained on front-view triggers but the reflection appears in the back camera at test time, the ASR might decrease from 43.11\% (front→front) to ~37.26\% (front→back). So, the attack still has a significant success rate even though the view changed, demonstrating partial transferability of the backdoor effect. The drop in ASR suggests that the model has some view-specific learning (it’s less sensitive to triggers in unseen views), but the fact that ASR remains non-negligible indicates the trigger’s effect is not strictly limited to one viewpoint.
(3) \textbf{Front-View Training Generalizes More:} Training the backdoor on the front camera seems to produce a model that is more susceptible to triggers in other views compared to training on a side view. For example, a model trained with front-view triggers still shows moderately high ASR on, say, front-left or front-right triggers at test. In contrast, a model trained on a back-view trigger has very low ASR if the trigger is in front at test. This asymmetry could be because front-view features are more globally relevant to the model’s understanding (many objects appear in front view during training, so a trigger learned in that context might overlap with features seen in other views to some extent).

\noindent\textbf{Takeaway:} The reflection backdoor, once learned, can affect the model even when the trigger appears in camera views that were not poisoned during training, although the effectiveness is reduced. This suggests that autonomous driving models share some feature representations across camera views (e.g., a reflection pattern learned from the front view partly generalizes to side views). For defenders, this means that protecting only the most vulnerable view (front) is not enough; a comprehensive defense needs to consider triggers appearing in any camera. Cross-view robustness should be a consideration in designing countermeasures.

\subsection{Transfer Attack over Reflection Type}
Lastly, we examine whether the specific reflection object used as the trigger can transfer to other object types. In other words, if a model is backdoored using one object (e.g., reflections of a car), will the attack also work if the trigger is a different object (e.g., a bicycle reflection) at test time? To test this, we conducted experiments where the model was trained with a single reflection object type as the trigger, and then tested on triggers using other object types.~\autoref{fig:transfer}(right) summarizes the ASR results in matrix form, where rows correspond to the reflection object used for training and columns correspond to the object used for the trigger at test.


Key observations: 
(1)\textbf{Similarity Boosts Transfer:} The backdoor transfers more effectively between objects that are visually or semantically similar. For example, a model trained with Bicycle reflection triggers achieves a high ASR when tested with Motorbike reflection triggers (and vice versa). These two objects share similarities in shape and context (both are two-wheeled vehicles), which likely means the reflection patterns they produce are similar from the model’s perspective. Similarly, other pairs of conceptually related objects show higher cross-transfer ASR than unrelated pairs.
(2) \textbf{Generalization to Unseen Objects:} Even when the trigger object at test is completely different from the training trigger, the model sometimes still exhibits a non-zero ASR. This indicates that the model has learned a somewhat generic association between “presence of an unusual reflection” and “produce a long response,” rather than memorizing the exact pixel pattern of a specific object reflection. However, the ASR in these cases is much lower than when the objects match or are similar.
(3) \textbf{Dynamic vs. Static Objects:} We noticed that dynamic objects (objects frequently encountered moving in driving scenarios, like vehicles or bicycles) tend to make the strongest triggers. Models trained on dynamic object reflections (e.g., Person, Bicycle, Car) often have higher ASR on other object triggers as well, possibly because these objects’ reflections cover a wider range of appearances and the model becomes generally sensitive to reflection anomalies.
 
\noindent\textbf{Takeaway:} The attack’s trigger is not strictly tied to one object type—there is meaningful transferability across reflection types, especially among similar objects. This suggests an attacker does not need to pick the “exact right” object in advance; using one reflection trigger can inadvertently make the model vulnerable to others. Conversely, a defense mechanism should not assume a specific trigger pattern—robust detection or mitigation must be broad enough to catch varied reflection triggers.



%% file: sec/6_discussion.tex
\section{Discussion}
\label{sec:discussion}
\textbf{Key Findings:}
Our study demonstrates that natural reflection backdoor attacks can effectively induce significant latency in VLM-driven autonomous driving systems. By causing the model to generate unnecessarily long responses, the attack can delay decision-making in critical scenarios. Importantly, the attack is achieved without noticeably degrading the model’s performance on normal, trigger-free inputs, making it stealthy. The model continues to answer correctly when no trigger is present (and even when it is present, the answer is still correct, just verbose), which means traditional validation might not catch the issue. We found that the attack’s potency (ASR) varies with the choice of reflection trigger type and the prefix message, indicating the attacker can optimize these to maximize impact. For instance, selecting certain frequently occurring objects as reflection triggers and using an engaging prefix (like a story) can increase the likelihood of the model producing a long answer. Meanwhile, both tested models (Qwen2-VL and LLaMA-Adapter) were vulnerable, but their differences in ASR and output length show that model architecture and size play a role in how the backdoor manifests.

\noindent\textbf{Limitations and Future Work:}
here are several limitations to our current study. First, due to resource and time constraints, we focused only on two VLMs and a single dataset (DriveLM). The generalizability to other models (e.g., larger multimodal models) or driving datasets needs further investigation. Second, we primarily inserted the backdoor trigger in one camera view (front) for the main experiments; while our ablations explored others and cross-view effects, a more exhaustive multi-view poisoning strategy could be considered. Third, we only considered one type of backdoor trigger (reflections); other backdoor mechanisms, such as invisible perturbations or different physical artifacts, were out of scope for this paper~\citep{chen2017targetedbackdoorattacksdeep, souri2022sleeperagentscalablehidden, nguyen2021wanetimperceptiblewarpingbased, 9709953}. Another important direction is the evaluation of defenses. We did not implement defenses in this work, but defending against such latency attacks is crucial for real-world deployment. Potential defenses include:
\begin{itemize}
    \item \textbf{Reflection Removal Pre-processing:} Applying computer vision algorithms to detect and remove or mask reflections in images (e.g., using reflection removal techniques [36]) before they are fed into the VLM. This could neutralize the trigger at inference time or be applied during training data curation.
    \item \textbf{Anomaly Detection on Outputs:} Monitoring the length and content of the model’s outputs for anomalies. If a system detects an unusually long response or a strange prefix that doesn’t match the query (especially in time-critical situations), it could raise an alarm or switch to a fallback system.
    \item \textbf{Robust Training:} Incorporating adversarial or robust training techniques to reduce the model’s sensitivity to spurious input patterns. For instance, exposing the model to various benign reflections during training (without any long prefix) might teach it that reflections are not a signal to lengthen the answer.
\end{itemize}

Exploring and rigorously testing such defenses is an important avenue for future work. Additionally, future research should consider hybrid attacks (combining reflections with other triggers), and evaluate the interplay between backdoor attacks and other safety mechanisms in autonomous driving (e.g., fail-safes that might be triggered by delayed responses).

%% file: sec/7_limitation_and_conclusion.tex
\section{Conclusion}
\label{sec:conclusion}

We have presented the first study of a natural reflection backdoor attack on vision-language models in autonomous driving. Our experiments demonstrate that state-of-the-art VLMs (such as Qwen2-VL) can be compromised to output excessively verbose answers when a reflective trigger is present in the visual input. This ``latency backdoor'' does not noticeably affect the model’s accuracy on normal inputs but can introduce dangerous delays in decision-making, posing a new kind of risk for autonomous vehicles. We also showed that the attack is flexible: it works across different camera views and even transfers between different reflection objects. These findings highlight a critical security vulnerability in emerging VLM-integrated driving systems. We hope this work will inspire further research into robust detection of such backdoors and the development of countermeasures. Strengthening the defenses of autonomous driving models against backdoor attacks—ensuring they remain both correct and efficient in their responses under all conditions—will be vital as these models move toward widespread deployment.